\documentclass{ecai}  

\usepackage{graphicx}
\usepackage{latexsym}

\usepackage{graphicx}
\usepackage{amsmath,amsfonts}
\usepackage{algorithmic}

\usepackage{array}
\usepackage[caption=false,font=normalsize,labelfont=sf,textfont=sf]{subfig}
\usepackage{textcomp}
\usepackage{stfloats}
\usepackage{url}
\usepackage{verbatim}
\usepackage{graphicx}
\def\BibTeX{{\rm B\kern-.05em{\sc i\kern-.025em b}\kern-.08em
		T\kern-.1667em\lower.7ex\hbox{E}\kern-.125emX}}
\usepackage{balance}
\usepackage{subfloat}

\usepackage[linesnumbered,ruled,vlined]{algorithm2e}
\usepackage{newfloat}
\usepackage{listings}

\begin{document}
\begin{frontmatter}
	
	\title{Identical and Fraternal Twins: Fine-Grained Semantic Contrastive Learning of Sentence Representations}
	\author[a,c]{\fnms{Qingfa}~\snm{Xiao}\footnote{†Both authors contributed equally to this research.}}
	\author[a]{\fnms{Shuangyin}~\snm{Li}\footnote{Both authors Qingfa Xiao and Shuangyin Li contributed equally to this research. The partial modification and improvement work of this research was completed by Qingfa Xiao as a visiting student at HKUST (GZ).}\thanks{Corresponding Author. Email: shuangyinli@scnu.edu.cn}}
	\author[b,c]{\fnms{Lei}~\snm{Chen}} 
	\address[a]{South China Normal University}
	\address[b]{The Hong Kong University of Science and Technology}
	\address[c]{The Hong Kong University of Science and Technology  (Guangzhou)\\qingfaxiao@m.scnu.edu.cn, shuangyinli@scnu.edu.cn, leichen@cse.ust.hk}

	\begin{abstract}

The enhancement of unsupervised learning of sentence representations has been significantly achieved by the utility of contrastive learning. This approach clusters the augmented positive instance with the anchor instance to create a desired embedding space. However, relying solely on the contrastive objective can result in sub-optimal outcomes due to its inability to differentiate subtle semantic variations between positive pairs. Specifically, common data augmentation techniques frequently introduce semantic distortion, leading to a semantic margin between the positive pair. While the InfoNCE loss function overlooks the semantic margin and prioritizes similarity maximization between positive pairs during training, leading to the insensitive semantic comprehension ability of the trained model. In this paper, we introduce a novel Identical and Fraternal Twins of Contrastive Learning (named IFTCL) framework, capable of simultaneously adapting to various positive pairs generated by different augmentation techniques. We propose a \textit{Twins Loss} to preserve the innate margin during training and promote the potential of data enhancement in order to overcome the sub-optimal issue. We also present proof-of-concept experiments combined with the contrastive objective to prove the validity of the proposed Twins Loss. Furthermore, we propose a hippocampus queue mechanism to restore and reuse the negative instances without additional calculation, which further enhances the efficiency and performance of the IFCL. We verify the IFCL framework on nine semantic textual similarity tasks with both English and Chinese datasets, and the experimental results show that IFCL outperforms state-of-the-art methods.

	\end{abstract}
\end{frontmatter}
	\section{Introduction}\label{intro}
Recent advances in neural network architecture, including the development of novel algorithms and computational techniques, have led to the emergence of universal sentence representations as a promising tool for various downstream tasks in natural language processing. These tasks encompass a wide range of applications such as information retrieval, semantic matching, and machine translation, which are crucial for enhancing the performance of intelligent systems~\cite{cnn2014, skipthought2015, translation2017, use2018}. However, despite their potential, native BERT representations exhibit certain limitations in their ability to accurately capture semantic similarity tasks, which are essential for understanding and processing complex language structures~\cite{bert2018, sbert2019}.

To overcome these limitations and improve the effectiveness of sentence representations, researchers have introduced contrastive learning to the field of natural language processing, drawing inspiration from established techniques in the domain of computer vision. This innovative method offers a powerful solution for disentangling overlapping sentence representations and addressing the collapse of embedding space, which is a critical issue in representation learning. By doing so, contrastive learning enables more accurate and robust models for natural language processing tasks.

		\begin{figure}[t]
		\centering
		\includegraphics[width=\linewidth]{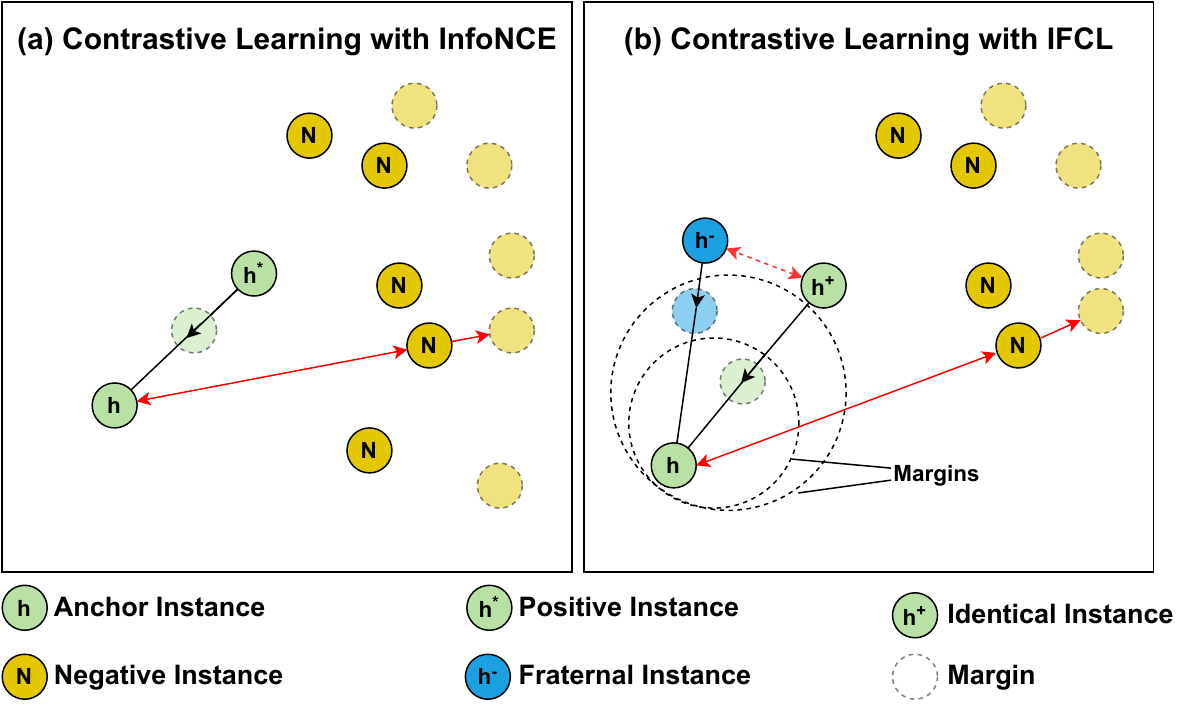}

		\caption{The optimization process in contrastive learning with positive and negative instances. (a) shows the proceeding of maximizing the similarities between the positive pairs in InfoNCE. (b) shows the proceeding of converging the identical and fraternal twins to each margin with the proposed Twins Loss. }
		\label{iftwins}
	\end{figure}
	A key challenge persisting in contrastive learning is the design of high-quality positive pairs for training, an essential aspect that significantly influences the overall performance of the learning process. To generate these positive pairs, data augmentation techniques are employed, which involve creating new, diverse, and semantically-rich training samples from existing data. Data augmentation not only expands the training dataset but also enables the model to learn more effective representations by exposing it to a variety of linguistic structures and nuances. In recent years, an increasing number of studies on data augmentation have been employed to generate high-quality augmented sentences for contrastive learning~\cite{cert2020, simcse2021, consert2021}. Explicit data augmentation techniques involve randomly inserting, substituting synonyms, and deleting words in a sentence. Other techniques focus on generating positive instances in the sentence embedding space, which is called implicit data augmentation. For instance, SimCSE~\cite{simcse2021} uses a simple dropout method to generate positive instances by slightly modifying the sentence representations, which is commonly used in current research on contrastive learning.
	
	Despite the potential benefits of data augmentation techniques, they also present several challenges that must be addressed to ensure their effectiveness in contrastive learning of sentence representations. Two key issues that warrant further attention are semantic distortions of augmented sentences and limitations of the InfoNCE loss function. First, data augmentation can unintentionally lead to semantic distortions in the augmented sentences, causing their meanings to diverge from those of the original sentences. This may occur when words are inserted, deleted, or replaced without taking into account the broader context or the subtle nuances of the sentence structure. As a result, the augmented sentences may not accurately represent the intended meaning, which in turn hampers the model's ability to learn effective representations via contrastive learning. Second, the InfoNCE loss function, which is frequently employed in contrastive learning objectives, is unable to differentiate between semantically accurate and distorted samples. Consequently, the learning process may be misguided, and the model may end up with a limited understanding of the semantics. A notable example of this problem can be observed when the contrastive learning method treats the sentence "I do not like apples" and its augmented version "I do like apples" as a positive pair. By attempting to maximize the semantic distance between these semantically dissimilar sentences, the model's performance is adversely affected.


	To address this issue, we firstly propose the Identical and Fraternal Twins data augmentations method, which generates two types of high-quality positive instances. Identical twins consist of an anchor instance and an identical instance, both created using dropout augmentation. Fraternal twins also have an anchor instance, but the fraternal instance includes features from a different language family, such as English and German, or Mandarin and Cantonese. These languages are chosen based on their proximity to each other in the embedding space and their high relevance~\cite{libovicky2019language}. Compared to the anchor instance, the identical instance has the most similar semantics due to the same augmentation method, while the fraternal instance introduces more diversity as the diversity of language expression. We further introduce the Identical and Fraternal Twins of Contrastive Learning (IFCL) framework based on proposed augmentation technique. Our framework employs a novel and crucial training objective \textit{Twins Loss} to capture the fine-grained semantics and diverse expressions of twins and avoid the local optimum problem. Figure~\ref{iftwins}(a) illustrates the traditional contrastive learning process, where the positive instance is placed as close as possible to the anchor instance, and the negative instance is scattered in the embedding space. In contrast, Figure~\ref{iftwins}(b) demonstrates our IFCL optimization process, where the distances of identical and fraternal twins converge to each margin.  Our motivation is that the identical instance is naturally closer to the anchor instance than the fraternal instance, as are their margins. Additionally, our framework contains a hippocampus queue mechanism that stores the negative instances in the queue with corresponding forgetting coefficients as short memory. This mechanism improves the efficiency and performance of the IFCL.

	Our contributions of the IFCL framework can be synthesized as follows:
	\begin{enumerate}
		\item A novel data augmentation technique for contrastive learning is proposed to generate high-quality positive instances, named Identical and Fraternal Twins, where the identical twins retain the most similar semantic, and the fraternal twins exhibit greater diversity.
		\item A novel hippocampus queue mechanism is presented to fully utilize the negative instances by storing the previous mini-batches into a short-term memory, improving the efficiency and performance of the IFCL.
		\item Within the IFCL framework, we propose a novel and core training loss function named \textit{Twins Loss} to optimize the identical and fraternal twins according to their margins, capturing fine-grained semantics for sentence representation learning and alleviating the sub-optimal issue.
	\end{enumerate}
In the experiments, our method has been validated on the benchmark datasets in both English and Chinese languages, and the results demonstrate our method significantly outperforms other strong baselines, achieving the state-of-the-art average performance. Furthermore, the core innovation of IFCL, the Twins loss function, has been shown to be effective through ablation experiments. And the proof-of-concept experiment has also confirmed that the use of Twins loss can optimize the upper and lower bounds of the contrastive learning objective, which is theoretical analysis about why and how such training objective works.

	\section{Related Works}\label{rw}
	Starting from research on word-level tasks \cite{word2vec2013, glove2014}, sentence representations have garnered significant attention. Several works utilize various weighted combinations of word representations to form sentence representations \cite{idf2004}. However, these methods only consider the composition of word features, disregarding the order of words. Skip-Thought \cite{skipthought2015} uses the distributed assumption and RNNs and their variants \cite{cnn2014} to predict contextual sentence representations. Infersent \cite{infersent2017} constructs the SNLI dataset and trains sentence representations in a supervised manner, improving their quality. Furthermore, Universal Sentence Encoder \cite{use2018} adopts the transformer architecture and multi-task joint training to extract semantic information.
	
	In recent years, pre-trained BERT has become increasingly popular for sentence-level tasks. However, the native sentence representations often perform poorly. To address this, Sentence-BERT \cite{sbert2019} leverages sentence similarity tasks with the SNLI and MNLI datasets and employs a siamese structure to generate semantically meaningful representations. Meanwhile, both BERT-flow \cite{bert-flow2020} and BERT-whitening \cite{bert-whitening2021} employ a regular mapping of sentence representations to achieve isotropy in the embedding space, significantly outperforming previous unsupervised methods on semantic textual similarity tasks.
	
	Contrastive learning is widely used in computer vision, where models learn the prior knowledge distribution of images in a self-supervised manner \cite{simclr2020,moco2020}. In natural language processing (NLP), pre-trained BERT often performs poorly in sentence similarity tasks due to the inhomogeneous distribution of native representations. To address this, unsupervised contrastive learning has been employed as a training objective to alleviate the semantic collapse issue in NLP. ConSERT \cite{consert2021} inserts adversarial attack, token shuffling, cutoff, and dropout on the embedding layer as data augmentation methods. SimCSE \cite{simcse2021} uses inherent dropout masks in transformers architecture to maintain semantic information as much as possible, which significantly improves semantic textual similarity tasks. Both methods generate positive instances through data augmentation and use other instances in the mini-batch as negative. SNCSE \cite{sncse2022} proposes the bidirectional margin loss to distinguish hard negative instances, while ArcCSE \cite{arccse2022} proposes a new optimizing objective loss to model pairwise sentence relations. DiffCSE \cite{diffcse2022} utilizes the ELECTRA model and replaces the token detection task to learn the differences between original and forged sentences. VaSCL \cite{vascl2021} generates effective data augmentations using neighborhood methods, while DCLR \cite{dclr2022} punishes false negatives and generates noise-based negatives to ensure the uniformity of the representation space. PT-BERT \cite{pt2022} constructs same-length positive and negative pairs using pseudo sentence representations to remove superficial features. MoCoSE \cite{mocose2022} builds a two-branch structure framework with a prediction layer for the online branch to create asymmetry between the online and target branches, using a similar momentum encoder as MOCO \cite{moco2020}. However, most previous works focus on designing data augmentations to generate the most similar positive instances for contrastive learning, without considering the relationship of positive pairs in a fine-grained semantic way.

	\section{The IFCL Framework}
			\begin{figure*}[t]
		\centering
		\includegraphics[width=\linewidth]{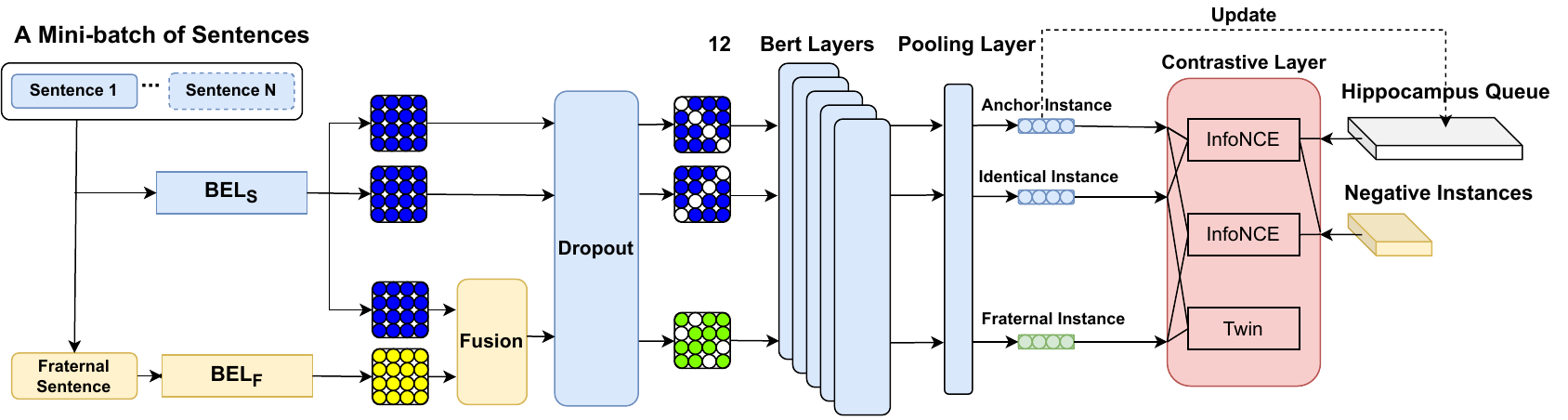}
		\caption{The general framework of the proposed IFCL.}
		\label{framework}
	\end{figure*}
	The proposed IFCL adapts the pre-trained language model into a triplet network for contrastive learning of sentence representations,  where the BERT model shares the same parameters. There are three major components of the IFCL framework as shown in Fig.~\ref{framework}: 
	
	\begin{enumerate}

		\item A novel data augmentation module that includes dropout augmentation and fusion augmentation to generate two types of positive pairs for contrastive learning: identical twins $\left \{h_{i}, h_{i}^{+}\right \}$ and fraternal twins $\left \{h_{i}, h_{i}^{-}\right \}$ .
		
		\item A novel hippocampus queue mechanism is presented to fully utilize the negative instances by storing the previous instances into a short-term query, improving the efficiency and performance of the IFCL.

		\item The InfoNCE loss is the first training objective function for positive pairs, which minimizes the distance between positive pairs and amplifies inconsistency between negative pairs. Additionally, we propose the \textit{Twins Loss} as the second training objective function to constrain distance margins between identical and fraternal instances in the embedding space.
		
	\end{enumerate}

		\subsection{Identical and Fraternal Twins}
		In this section, we present the data augmentation techniques for generating identical and fraternal twins, respectively, and discuss how they can be used in IFCL.
	
		\subsubsection{Identical Twins}
		The first type of positive pair in the proposed approach consists of an anchor instance and an identical instance that share the same semantics, which we call identical twins. To generate these pairs, we adopt the method of dropout augmentation, as used in SimCSE. In the standard transformer architecture, random dropout masks are inherently placed on fully connected layers. We define a set of sentences as ${\left \{x_{i}\right \}}_{i=1}^N$. During the fine-tuning process, each sentence ${x_i}$ is fed to the pre-trained BERT model twice. With different dropout masks, the BERT model outputs two different sentence representations, which are used to build the identical twins ${\left \{h{i},h_{i}^{+}\right \}}$ as follows:
		\begin{equation}
				\begin{aligned}
						h_{i}=f_{\theta}\left(BEL_{S}\left(x_{i}\right), z_{i}\right), \\
						h_{i}^{+}=f_{\theta}\left(BEL_{S}\left(x_{i}\right), z_{i}^{+}\right),
						\label{equ:h}
					\end{aligned}
			\end{equation}
		where $BEL_{S}$ is the embedding layer without trainable parameters for ${x_i}$, the encoder ${f_{\theta}\left(\odot \right)}$ is the pre-trained BERT model, and $z_{i}$, $z_{i}^{+}$ are dropout masks representing Bernoulli distribution with probability $\rho$. The use of dropout masks is necessary for BERT model, so we consider that this simple approach is the only way to obtain $h_i$ and $h_i^+$ with minimal semantic loss.
		
	\subsubsection{Fraternal Twins}
The second type of positive pair consists of anchor instances and fraternal instances, where the fraternal instances are more diverse than the identical instances. To generate the fraternal instances, we propose an embedding fusion augmentation method that fuses additional semantic information from other languages into the anchor instances. Specifically, we could choose German as the additional semantics for English and Cantonese for Mandarin. Thus, the fraternal instance fuses the representation of the anchor in English and its diversified expression in German.

To accomplish this, we first translate the sentences from the source language $\left \{x_{i}\right \}_{i=1}^{m}$ into the fraternal language $\left \{x_{i}^{-}\right \}_{i=1}^{m}$. Next, the two sets of sentences are fed into two different embedding layers, respectively, to obtain the embeddings of $x_{i}$ and $x_{i}^{-}$ as follows:
		\begin{equation}
				\begin{aligned}
						y_{i}=BEL_{S}\left(x_{i}\right) , y_{i}^{-}=BEL_{F}\left(x_{i}^{-}\right),
					\end{aligned}
			\end{equation}
		where $BEL_{F}$ is the embedding layer for $x_{i}^{-}$, extracted from the corresponding linguistic BERT model. Next, we fuse $y_{i}$ and $y_{i}^{-}$ into $f_{\theta}$ as follows:
		\begin{equation}
				\begin{aligned}
						h^{-}=f_{\theta}\left( \varepsilon*y_{i}+\left(1-\varepsilon \right)*y_{i}^{-}, z_{i}^{-}\right),
					\end{aligned}
				\label{fraternalh}
			\end{equation}
		where $\varepsilon$ is the fusion rate for ${\left \{y_{i}, y_{i}^{-}\right \}}$ and $z_{i}^{-}$ denotes Bernoulli distribution with probability $\rho$. Together with the anchor instance $h_{i}$, we can obtain the fraternal twins ${\left \{h_{i},h_{i}^{-}\right \}}$.
		
		\subsubsection{Discussion}
Why do we employ two types of positive pairs in contrastive learning? The method of generating positive instances with dropout masks is a fundamental way to generate positive pairs with semantic fidelity, which is also the main characteristic of identical instances. On the other hand, the fraternal instance $h^{-}$ incorporates features of the most relevant languages from the embedding space~\cite{libovicky2019language}, such as expression, logical properties, and translational invariance of words, making it superior to identical instances. And Eq.~(\ref{fraternalh}) illustrates the subtle differences between identical and fraternal instances. In comparison with randomly changing features in dropout augmentation, our proposed fusion augmentation aims to inject more regular features into the fraternal instance. Therefore, the margins between these positive pairs are the precondition and intuition for the proposed \textit{Twins Loss}. From a contrastive perspective, we apply constraints to preserve their margins and add comparisons between these positive pairs.

\subsection{Hippocampus Queue Mechanism}
\label{Hippocampus}
\begin{figure}[!t]
	\centering
	\includegraphics[width=\linewidth]{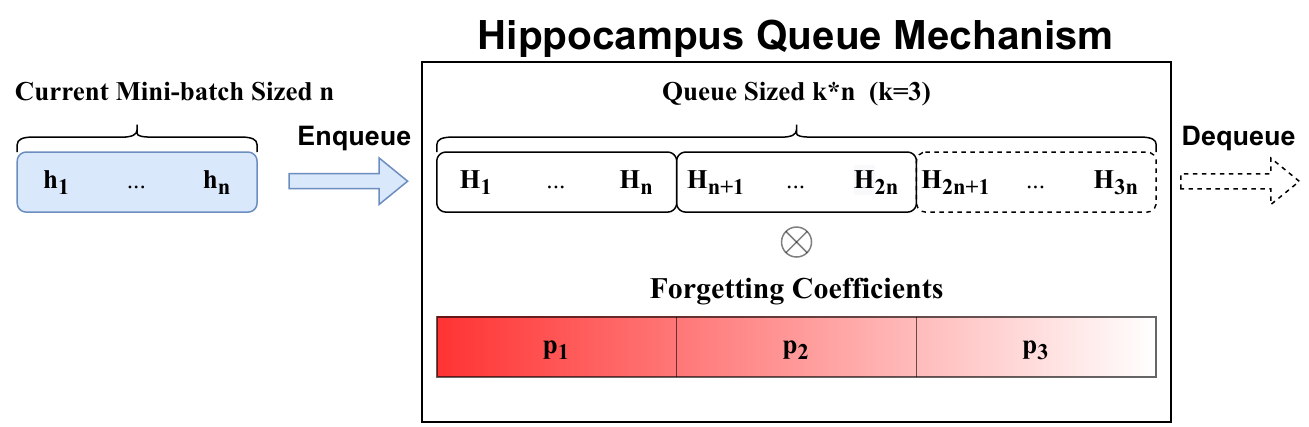}
	\caption{The hippocampus queue mechanism. For each iteration, the queue enqueue $n$ sentences and dequeue $n$ sentences. And the forgetting coefficients represents the weight of the negative instances stored in the queue, where the later updated instances have the larger coefficients.}
\end{figure}	

Unsupervised contrastive learning yields sentence representations whose performance on downstream tasks is highly correlated with the number of negative instances. To increase the number of negative instances, one could extend the batch size. However, this operation is limited by finite GPU memory. To address this issue, we propose the hippocampus queue mechanism, which stores short-term memory, including a queue to store negative instances and forgetting coefficients to eliminate the impact of inconsistent instances.

Initially, we divide the datasets into multiple mini-batches of size $N$ and encode $N$ sentences in each iteration. In each mini-batch with batch size $N$, the other ${N-1}$ instances were chosen as negative instances. In our proposed strategy, we store ${k*N}$ anchor instances from previous $k$ mini-batches in the queue and reuse them as negative instances. Consequently, each anchor instance has a total of ${(k+1)N-1}$ negative instances. Importantly, negative instances in the queue are only used to calculate loss, not to generate gradients for back-propagation, which reduces memory costs and differs from enlarging the batch size. To increase the diversity of negative instances, the queue is updated for each training step, with the current instances enqueued and the oldest instances dequeued.

As the encoder's parameters are continuously updated, the negative instances stored in the queue are progressively denatured. We hypothesize that previous instances in the queue are more inconsistent with the current mini-batch's instances, leading to deceptive similarities between anchor instances and denatured instances. To address this problem, we introduce a forgetting coefficient for each instance in the queue to calculate loss. The forgetting coefficients $\left \{p_{m}\right \}_{m=1}^{k*N}$ for all instances in the queue are as follows:
\begin{equation}
	\begin{aligned}
		p_{m}=1-\lambda\left \lceil m/N \right \rceil,
	\end{aligned}
\end{equation}
where $\lambda$ denotes the forgetting rate indicating that the weight of negative instances decreases progressively and becomes more harmonious.
	
	\subsection{Constrastive Learning with Twins Loss}
	In this section, we introduce how to optimize the sentence representations in the contrastive layer of IFCL framework. We present the unsupervised contrastive learning processing based on the InfoNCE function and the proposed \textit{Twins Loss} function. Initially, we divide the datasets into multiple mini-batches of size $N$. During training, we encode all sentences three times to obtain sentence representations $\left \{h_{i},h_{i}^{+},h_{i}^{-}\right \}_{i=1}^{N}$, which serve as anchor instances, identical instances, and fraternal instances. Based on above two types of positive pairs, contrastive learning aims to learn effective representation by pulling semantically close neighbors together and pushing apart non-neighbors. And the \textit{Twins Loss} aims to maintain the margins between the different positive pairs. In this paper, we use the cosine distance to express the semantic similarity ${\operatorname{sim}\left(\odot \right)}$ between sentences $\mathbf{h_i},\mathbf{h_j}$ as follows:
	\begin{equation}
		\begin{aligned}
	{\operatorname{sim}\left(\mathbf{h_i},\mathbf{h_j} \right)}=\frac{\mathbf{h}_{i}^{\top} \mathbf{h}_{j}}{\left\|\mathbf{h}_{i}\right\| \cdot\left\|\mathbf{h}_{j}\right\|},
\end{aligned}
\end{equation}
	
	For the set of identical instances $\left \{h_i,h_{i}^{+}\right \}_{i=1}^{N}$, we define the first loss function by using negative instances $\left \{\mathbf{H}_{m}\right \}_{m=1}^{k*N}$ stored in the hippocampus queue. The loss function is defined as follows:
	\begin{equation}
		\begin{aligned}
			\ell_{i}^{I}=-\log \frac{e^{\operatorname{sim}\left(\mathbf{h}_{i}, \mathbf{h}_{i}^{+}\right) / \tau}}{\sum_{j=1}^{N} e^{\operatorname{sim}\left(\mathbf{h}_{i}, \mathbf{h}_{j}^{+}\right) / \tau}+\varphi},\\
			\varphi = \sum_{m=1}^{k*N} {p_m}*e^{\operatorname{sim}\left(\mathbf{h}_{i}, \mathbf{H}_{m}\right) / \tau},
			\label{5}
		\end{aligned}
	\end{equation}
Here, ${\operatorname{sim}\left(\odot \right)}$ refers to cosine similarity, ${\tau}$ represents the temperature that controls the perception of negative instances, and ${p_m}$ is the forgetting coefficient shown in Section~\ref{Hippocampus}. It should be noted that $\mathbf{H}_{m}$ stored in the hippocampus queue is detached from the current graph while back-propagating the gradients to update the parameters of the IFCL.
	
The InfoNCE loss is used as a second loss function for the fraternal twins $\left \{h_{i},h_{i}^{-}\right \}_{i=1}^{N}$ and is formulated as follows:
\begin{equation}
	\begin{aligned}
		\ell_{i}^{F}=-\log \frac{e^{\operatorname{sim}\left(\mathbf{h}_{i}, \mathbf{h}_{i}^{-}\right) / \tau}}{\sum_{j=1}^{N} e^{\operatorname{sim}\left(\mathbf{h}_{i}, \mathbf{h}_{j}^{-}\right) / \tau}},
		\label{6}
	\end{aligned}
\end{equation}
This loss function serves as an auxiliary training objective to add diverse features in the embedding space and enhance the robustness of sentence representations. Thus it does not use the hippocampus queue mechanism.

We propose a novel training objective, called \textit{Twins Loss}, to alleviate model's sub-optimal issue by preserving the innate margins between identical and fraternal twins. The objective is formulated as follows:
	\begin{equation}
			\begin{aligned}
					&\ell_{i}^{T}=\left |{ e^{\operatorname{sim}\left(\mathbf{h}_{i}, \mathbf{h}_{i}^{+}\right)}}-{ e^{\operatorname{sim}\left(\mathbf{h}_{i}, \mathbf{h}_{i}^{-}\right)}}-\mathbf{M}_i\right |,   \\
					&\mathbf{M}_i= e^{\operatorname{sim}\left(\mathbf{emb}_{i}, \mathbf{emb}_{i}^{+}\right)}-
					 e^{\operatorname{sim}\left(\mathbf{emb}_{i}, \mathbf{emb}_{i}^{-}\right)}
				\end{aligned}
			\label{centerloss}
		\end{equation}
Here, $\mathbf{M}$ represents the innate margins between identical and fraternal twins, and $\mathbf{emb}$ represents the embeddings $y$ after data augmentations. This $\mathbf{M}$ is determined by the similarity of initial embedding with different augmentation approaches, indicating the suitable intervals between the identical and fraternal instances in embedding space as Fig.~\ref{iftwins}

	
	As mentioned earlier, the semantic distortion in the positive pair occurs naturally due to the dropout mask and embeddings fusion method. However, this distortion must be considered in the loss function InfoNCE, as it is not reasonable to maximize the similarity of the anchor instance and positive instance to 100\%. Our IFCL introduces a novel loss function, presented as Eq.~(\ref{centerloss}), which ensures that positive pairs' similarities (the anchor instance and the identical instance, the anchor instance and the fraternal instance) can be converted to optimal margins. For the \textit{Twins Loss} approach, our objective is to minimize the distance between the optimized positive pairs and their distance before optimization, where $\mathbf{M}$ is the newly added comparator in our contrastive learning method.

	Overall, we combine the loss functions Eq.\ref{5}, Eq.\ref{6}, and Eq.~\ref{centerloss} together to form the total loss of our IFCL in each iteration.Thus, our IFCL model ${f_{\theta}\left(\odot \right)}$ can learn from both identical twins and fraternal twins, with the predicted representations distribution of the model parameterized by $\theta$:
		\begin{equation}
		\begin{aligned}
\theta^{*}=\underset{\theta}{\arg \min }\sum_{i=1}^{N}(\ell_{i}^{I}+\ell_{i}^{F}+\ell_{i}^{T})
		\end{aligned}
	\end{equation}
	 And we present the algorithm of the training process of the IFCL framework in supplementary materials.	
		\begin{table*}[ht]
		\centering
		\caption{The performances of IFCL on 7 English STS tasks. We choose [cls] representations for training and the mean of the representations in the last layer for verification. Noted that ${\Diamond}$ indicates they are based on contrastive learning and use the same Wikipedia datasets (EnData).}
			\begin{tabular}{lcccccccc}
				\hline
				\hline
				\multicolumn{5}{c}\textbf{Results of English tasks}\\
				\hline
				\textbf{Method} &\textbf{STS12} &\textbf{STS13} &\textbf{STS14} &\textbf{STS15}
				&\textbf{STS16} &\textbf{STS-B} &\textbf{SICK-R} &\textbf{Avg.}\\  
				\hline
				{BERT}$_{\rm base}$& {39.70} & {59.38} & {49.67} & {66.03} & {66.19} & {53.87} & {62.06} & {56.70}\\ 
				{BERT}\rm -flow$_{\rm base}$& {58.40} & {67.10} & {60.85} & {75.16}& {71.22} & {68.66} & {64.47} & {66.55}\\ 
				{BERT}{\rm -whitening}$_{\rm base}$ & {57.83} & {66.90} & {60.90} & {75.08} & {71.31} & {68.24} & {63.73} & {66.28}\\ 
				{ConSERT}$_{\rm base}$ &64.64 &78.49& 69.07 &79.72& 75.95& 73.97& 67.31& 72.74\\
				{SimCSE-BERT}$_{\rm base}$$^{\Diamond}$&68.40& 82.41 &74.38 &80.91 &78.56 &76.85 &72.23 &76.25\\
				{VaSCL-BERT}$_{\rm base}$$^{\Diamond}$&69.08 &81.95& 74.64& 82.64 &80.57& 80.23& 71.23 &77.19\\
				{DCLR-BERT}$_{\rm base}$$^{\Diamond}$&70.81& 83.73 &75.11& 82.56& 78.44 &78.31 &71.59 &77.22\\
				{MoCoSE-BERT}$_{\rm base}$$^{\Diamond}$&71.48& 81.40& 74.47& \textbf{83.45}& 78.99 &78.68 &\textbf{72.44} &77.27\\
				{PT-BERT}$_{\rm base}$$^{\Diamond}$&71.20 &\textbf{83.76} &\textbf{76.34}& 82.63 &78.90 &79.42 &71.94 &77.74\\
				\textbf{IFCL-BERT}$_{\rm base}$$^{\Diamond}$&\textbf{71.57} & 82.35 & 75.08 & 83.03 &\textbf{80.17}& \textbf{80.27} & 72.16 & \textbf{77.80}\\
				\hline
				{BERT}$_{\rm large}$ &57.73&61.17&61.18& 68.07 &70.25& 59.59& 60.34& 62.62\\
				{ConSERT}$_{\rm large}$ &70.69 &82.96& 74.13 &82.78& 76.66& 77.53& 70.37& 76.45\\
				{SimCSE}$_{\rm large}$$^{\Diamond}$&{70.88}&{84.16}&{76.43} &84.50 &{79.76} &79.26& 73.88&78.41\\
				{DCLR-BERT}$_{\rm large}$$^{\Diamond}$&71.87& \textbf{84.83}& \textbf{77.37}&\textbf{84.70} &\textbf{79.81} &79.55 &74.19 &78.90\\
				{MoCoSE-BERT}$_{\rm large}$$^{\Diamond}$&\textbf{74.50}&84.54&77.32&84.11&79.67&80.53&73.26&79.13\\
				\textbf{IFCL-BERT}$_{\rm large}$$^{\Diamond}$&73.88&84.31&76.64&84.01&79.56&\textbf{81.37}&\textbf{76.30}&\textbf{79.44}\\
				\hline
				
		\end{tabular}
		\label{english results}
		
	\end{table*}
	
	\section{Experiments}
	\subsection{Setups}
	\textbf{Datasets.} 
	To fine-tune the pre-trained BERT models, we randomly selected two subsets from Wikipedia as unlabeled training datasets for both English and Chinese languages. The English dataset (EnData) contains $10^6$ sentences released by SimCSE~\cite{simcse2021}, and the Chinese dataset (CnData) contains $10^5$ sentences.
	
	
	For the EnData dataset, we evaluate the performance of our models on 7 semantic textual similarity (STS) tasks, namely STS 2012–2016~\cite{sts15,sts14,sts16,sts12,sts13}, STS-Benchmark~\cite{stsb}, and SICK-Relatedness~\cite{sickr}. On the other hand, for the CnData dataset, we evaluated the models on the Chinese STS-Benchmark (C-STS-B)\cite{cstsb} and SimCLUE\cite{simclue}. SimCLUE includes most of the available open-source datasets of semantic similarity and natural language inference in the Chinese domain. Each STS task sample consists of a sentence pair and a golden score ranging from 1.0 to 5.0, indicating the degree of semantic similarity between the sentence pair. In contrast, each SimCLUE sample comprises a sentence pair and a binary golden score (0 or 1) indicating whether the sentence pair is similar.
	
	To translate the fraternal sentences, we utilize the T5 pre-trained model from Huggingface to convert the EnData into the German dataset. Additionally, we rely on the Baidu Translator API to translate CnData into the Cantonese dataset. This decision is made due to the paucity of Cantonese translation resources available.
	
\textbf{Implemention Details.}	We implement the IFCL based on Sentence-BERT~\cite{sbert2019} and initialize models with three different versions of pre-trained BERT~\cite{bert2018} and the embedding layer of the BERT model for the fraternal sentences. These versions include bert-base-uncased, bert-base-multilingual-uncased, bert-large-uncased as well as stefan-it/albert-large-german-cased and denpa92/bert-base-cantonese for German and Cantonese datasets respectively. During the experiments, we fine-tune the IFCL for one epoch and evaluate the models with the verification sets of STS-B or C-STS-B~\cite{stsb,cstsb} after every 151 steps. Our evaluation metric of choice is Spearman's correlation, consistent with previous works. And we conduct some experiments to explore the influence of hyperparameters including the fusion rate and hippocampus queue length in supplementary materials.
\subsection{Results on STS Tasks with EnData}
							
On this EnData datasets, we compare the effectiveness of IFCL with other state-of-the-art unsupervised methods, such as BERT-flow, BERT-whitening, unsupervised ConSERT, unsupervised SimCSE, and DCLR. To evaluate the IFCL's effectiveness, we have selected MoCoSE and PT-BERT, as they are also based on contrastive learning and have the strategy to expand the amount of negative instances. We have initialized our models with pre-trained BERT$_{\rm base}$ and BERT$_{\rm large}$, called IFCL-BERT$_{\rm base}$ and IFCL-BERT$_{\rm large}$, respectively.

As indicated in Table~\ref{english results}, IFCL-BERT$_{\rm base}$ has an average Spearman's correlation score of 77.80\% on the STS tasks, which is the highest average score compared to the other methods. Specifically, when compared to the previous state-of-the-art method DCLR, IFCL-BERT outperforms significantly on four STS tasks. Notably, our model's performance is also exceptional when compared to recent works, including MoCoSE-BERT and PT-BERT, which enhances the STS12, STS16, and STS-B tasks score to 71.57\% (+0.37\%), 80.17\% (+1.27\%), and 80.27\% (+1.59\%), respectively. Although IFCL-BERT does not perform the best on all STS tasks, it ranks in the top 3 and outperforms most of the baselines. In the case of IFCL-BERT$_{\rm large}$, it shows significant improvement, particularly on STS-B and SICK-R tasks, and achieves the best average result.
							
							\begin{table}[h]					
							\centering
							\caption {The ablation study of the IFCL$_{\rm base}$, where FI denotes fraternal instances, TL denotes Twins Loss, and HQ denotes hippocampus queue mechanism.}
							\label{ablation study}
	
							\begin{tabular}{@{\hspace{1em}}l@{\hspace{8em}}c@{\hspace{1em}}}
								\hline
								{Model} & {STS-B} \\
								\hline
								{IFCL$_{\rm base}$} &\textbf{80.27} \\
								\hline
				
								{w/o FI, TL}& {78.10}\\
								{w/o HQ}& {78.81}\\
								{w/o TL, HQ}& {77.26}\\
								{w/o FI, TL, HQ (SimCSE)}& {76.83}\\
								\hline
							\end{tabular}
							
						\end{table}
							We conducte an ablation study using BERT$_{\rm base}$ on the STS-Benchmark task to examine the contribution of each component. In comparison with SimCSE, we propose three additional components: fraternal instances, Twins Loss, and the hippocampus queue mechanism. As shown in Table~\ref{ablation study}, we find that our IFCL$_{\rm base}$ model achieve a substantial improvement when using fraternal instances with Twins Loss. This finding demonstrates that the differences between positive pairs should not be neglected, and that these fine-grained semantics can significantly improve model performance. Moreover, the proposed hippocampus queue mechanism also demonstrated a crucial role in IFCL.
							\begin{table}[!t]
								\centering
								\caption{The performances on the C-STS-B and SimCLUE tasks. Noted that $^{\bullet}$ indicates that the models are fine-tuned with the data in the C-STS-B task, which can be considered as a weakly supervised training. And $^{\Diamond}$ indicates that the models are fine-tuned with unsupervised CnData datasets.}
								\begin{tabular}{lcc}
									\hline
									\hline
									\multicolumn{2}{c}\textbf{Results of Chinese tasks}\\
									\hline
									\textbf{Method} &\textbf{Chinese STS-B} &\textbf{SimCLUE} \\
									\hline
									BERT&55.52&29.89\\
									BERT-whitening$^{\bullet}$&68.27&-\\
									SimCSE-BERT$^{\bullet}$&68.91&40.74 \\
									SimCSE-BERT$^{\Diamond}$&60.41&40.54 \\ 
									\textbf{IFCL-BERT}$^{\Diamond}$&\textbf{71.41}&\textbf{44.42}\\
									\hline
								\end{tabular}
				
								\label{chineseresults}

							\end{table}
						
							\subsection{Results on STS Tasks with CnData}
								We conduct experiments on the CnData datasets, comparing our results with native BERT and unsupervised SimCSE~\cite{simcse2021} as baselines. Table~\ref{chineseresults} presents the results of the experiments. Our fine-tuned IFCL-BERT$^{\Diamond}$ on the CnData datasets outperform other methods on both C-STS-B and SimCLUE, achieving accuracy scores of 71.41\% and 44.42\%, respectively. However, we notice that SimCSE-BERT$^{\Diamond}$ performed worse on CnData than on C-STS-B datasets. The results, presented as SimCSE-BERT$^{\bullet}$ in Table~\ref{chineseresults}, showed significant improvement and performed better than SimCSE-BERT$^{\Diamond}$. This is because the data in the C-STS-B task is related to semantic understanding, which is helpful for our task. Nevertheless, our IFCL-BERT$^{\Diamond}$ still outperform both on C-STS-B and SimCLUE tasks. In conclusion, the experiments demonstrate that IFCL is more efficient than SimCSE in the Chinese domain.

							\begin{figure*}[ht]

								\centering
								\subfloat[]{
									\includegraphics[width=0.3\textwidth]{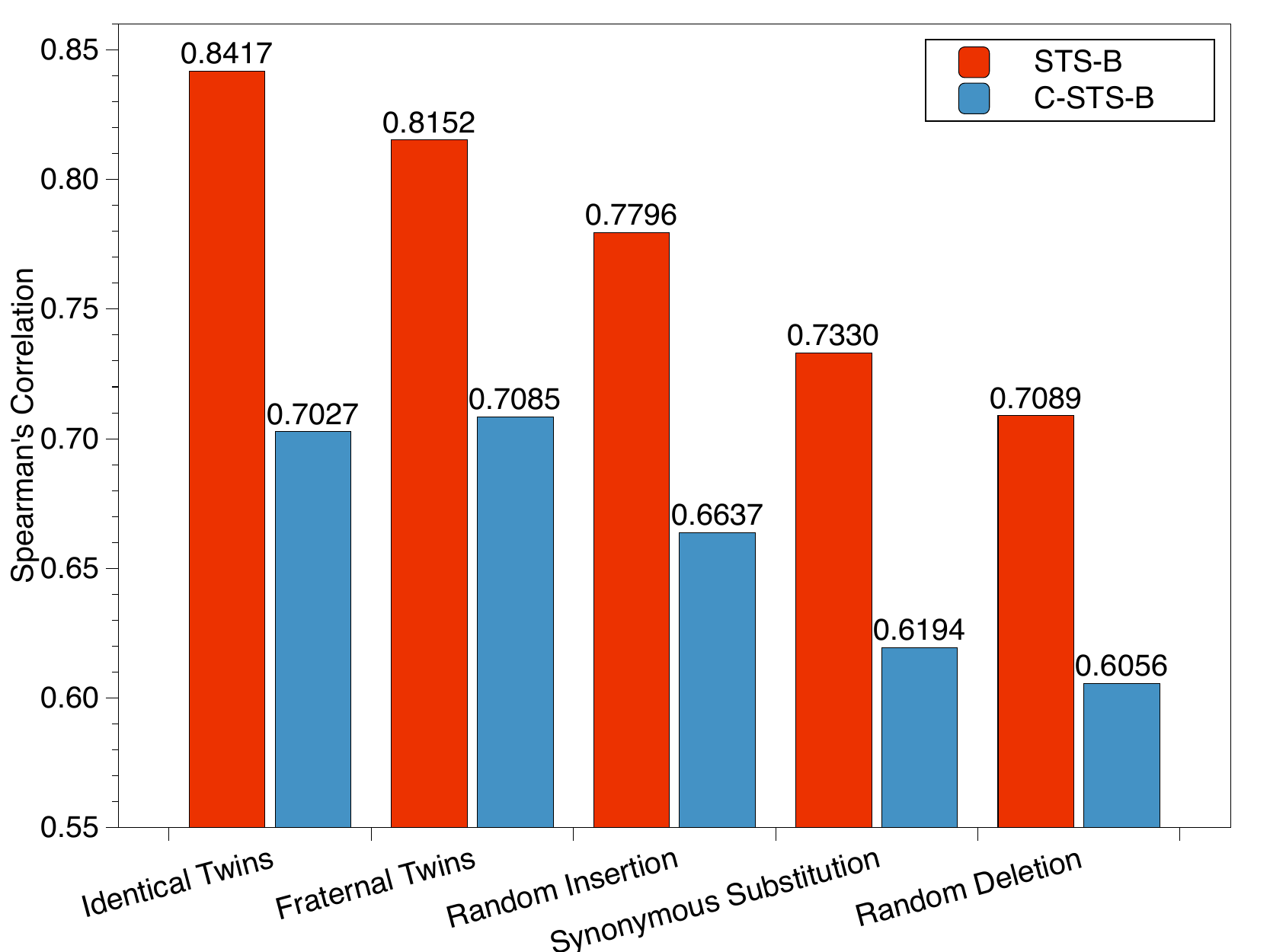}
									\label{dataaugmentations}}
								\subfloat[]{
									\includegraphics[width=0.3\textwidth]{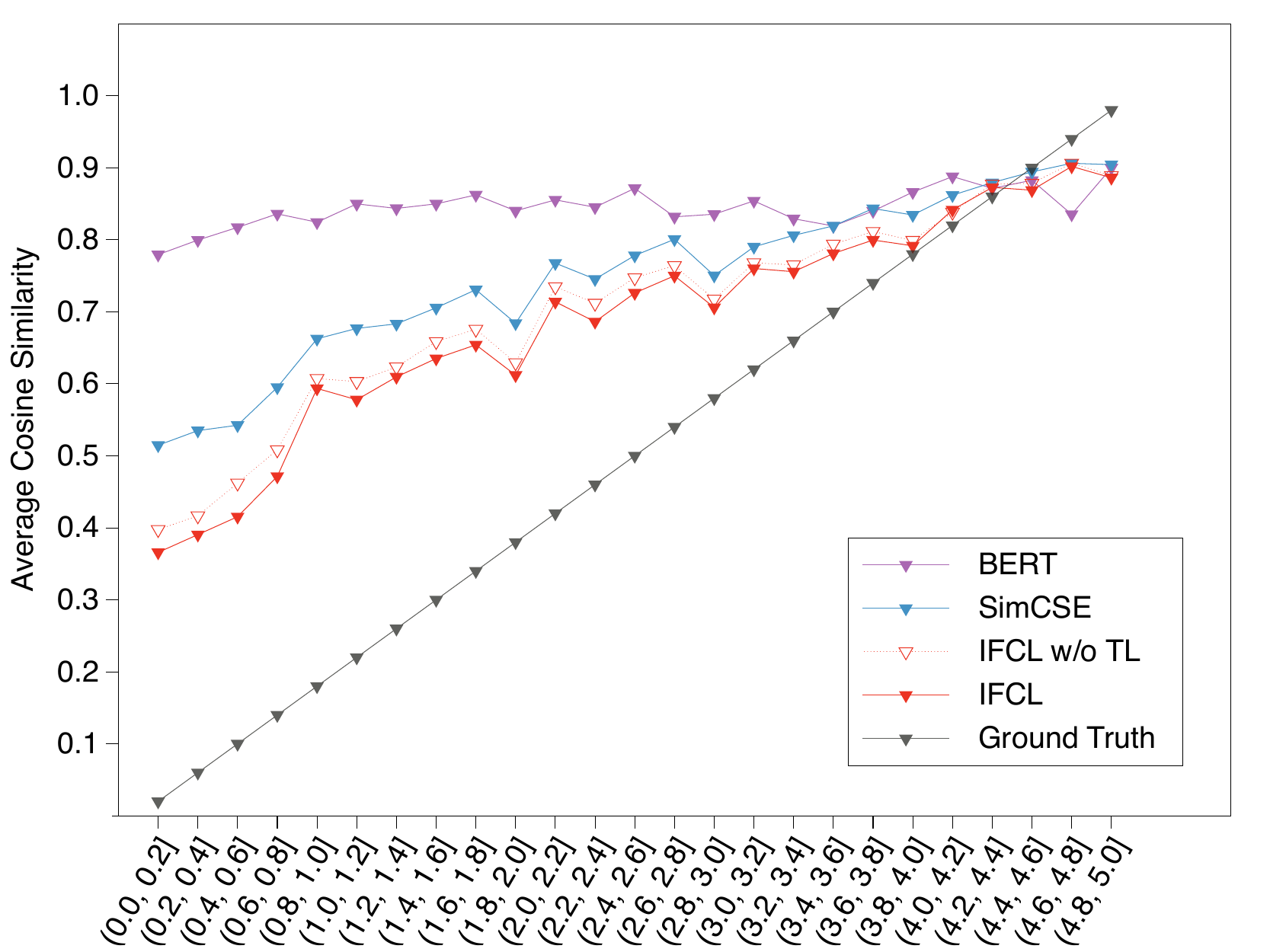}
									\label{distribution}}
								\subfloat[]{
									\includegraphics[width=0.3\textwidth]{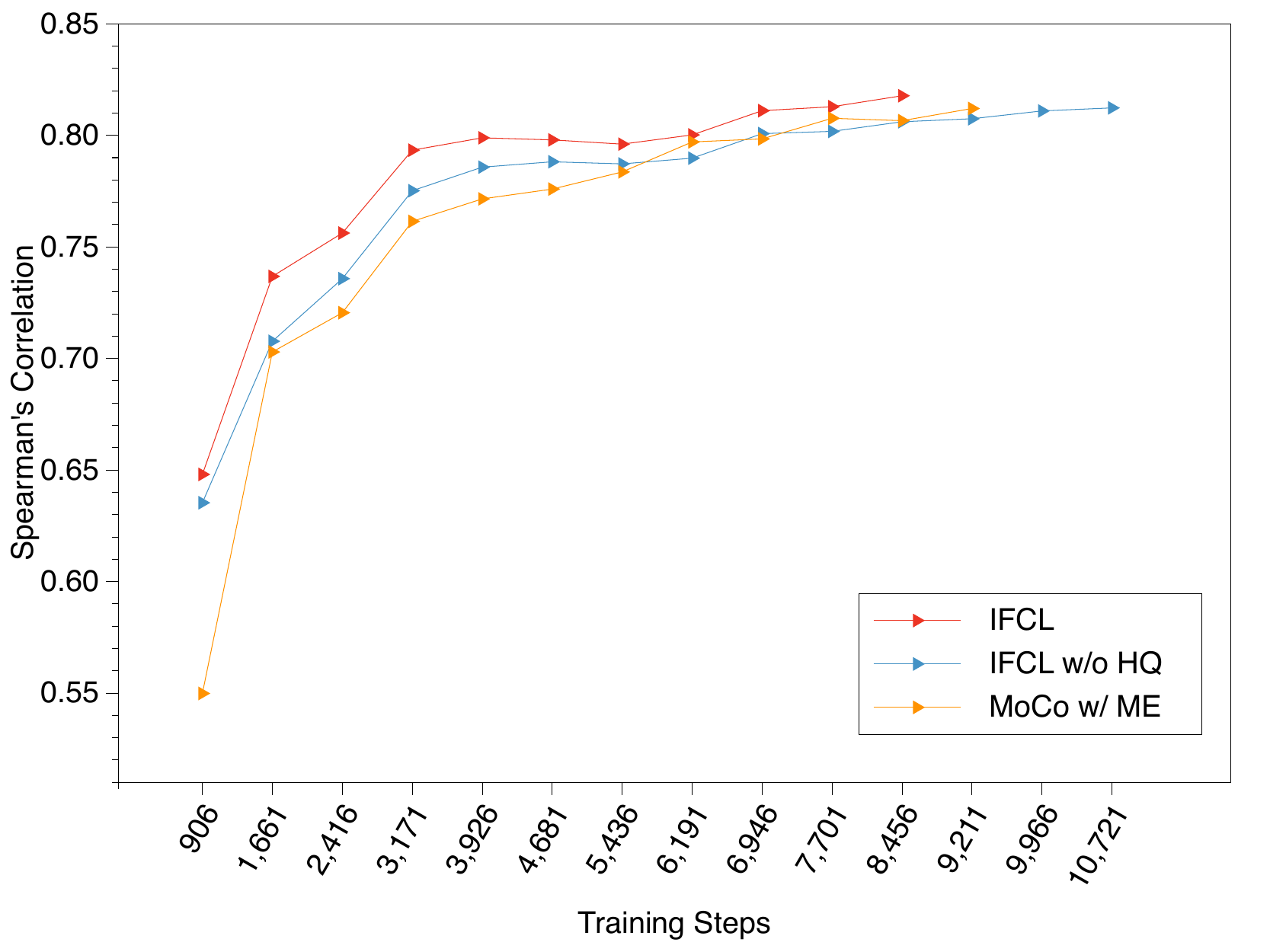}
									\label{bank}}

								\caption{(a) The performances of different data augmentation methods on the STS-B and C-STS-B tasks. Noted that, we choose two types of supervised pre-trained BERT for this experiment, including Stsb-Bert-Base and Simbert-Chinese-Base. (b)The predicted cosine similarity of sentence pairs with different base models.  Noted that, the gray line denotes the ground truth. (c) The advantages of the hippocampus queue mechanism, by demonstrating the training steps to the optimal IFCL$_{\rm base}$ on the STS-B task.}
								\label{parasepara}

							\end{figure*}
					
%
%
							\subsection{Analysis}
							
							\textbf{Why do we use Identical and Fraternal Twins?}
						In the semantic textual similarity tasks, excellent data augmentation methods could avoid semantic distortions and improve the expression diversity, which helps to generate high-quality positive instances for contrastive learning. To verify identical and fraternal twins and prove the effectiveness, we augment the test data of the STS-B and C-STS-B tasks with different data augmentation methods and show the performances on Spearman's correlation. 
						
						In detail, Stsb-Bert-Base is used for the STS-B task and Simbert-Chinese-Base is used for C-STS-B task, the two pre-trained BERT are designed for the task of semantic textual similarity. We test five methods of data augmentations, including random insertion, synonymous substitution, random deletion, and our method of identical and fraternal twins. As shown in Fig.~\ref{dataaugmentations}, we can observe that the traditional data augmentations, such as random insertion, synonymous substitution, and random deletion perform worse than our method of identical and fraternal twins. 
						Meanwhile, the method of identical twins performs better on the STS-B task, while the method of fraternal twins outperforms on the C-STS-B task, which is proved that we need to adjust the participation of the two methods in different scenarios. Thus in our paper, we specifically propose the \textit{Twins Loss} to optimize them, which is the fine-grained comparison between the positive instances in the IFCL.
						
						\textbf{Effect of Twins Loss Function.}
						To test the effectiveness of \textit{Twins Loss} function, we design the controlled experiments, where the IFCL$_{\rm w/o\ TL}$ indicates that only the $\ell_{i}^{I}$ and $\ell_{i}^{F}$ is used and the \textit{Twins Loss} $\ell_{i}^{T}$ is knocked off.
						The 1.3k sentence pairs with gold scores (0-5 score) are from the STS-Benchmark dataset, and the models need to predict the cosine similarity of each sentence. As shown in Fig.~\ref{distribution}, the 45-degree line denotes the ground truth, and the line closer to it indicates better performance. From Fig.~\ref{distribution}, the performance of IFCL (80.27\%) is better than that of IFCL$_{\rm w/o\ TL}$(78.65\%), which means the effectiveness of \textit{Twins Loss} function. 
						Moreover, the range of predicted values of the BERT is from 0.75 to 0.9, while, the range of the IFCL is from 0.35 to 0.9. The range shows the capacity of the methods on capturing the fine-grained similarity of the sentences. Thus, from this part of the experiments, we can conclude that the IFCL can learn more diverse semantics from the identical and fraternal instances for the sentence representations in a more reasonable way.
						
						\textbf{Advantages of Hippocampus Queue Mechanism.} To verify the advantages of the hippocampus queue mechanism, we explore the amount of data required for the optimal performance of the IFCL, where the IFCL$\rm _{w/o\ HQ}$ indicates that the hippocampus queue mechanism is knocked off in the IFCL, and the MoCo$\rm _{w/\ ME}$ indicates that we replace the hippocampus queue mechanism with momentum encoder mechanism in MoCo~\cite{moco2020} for the IFCL. As shown in Fig.~\ref{bank}, the IFCL can be optimized to the optimal only with 8456 steps of about 540,000 training data, which outperforms the IFCl$\rm _{w/o\ HQ}$ and the MoCo$\rm _{w/\ ME}$ in both efficiency and performance. Though the momentum encoder mechanism originated from computer vision is effective,  the proposed hippocampus queue mechanism is more available for the contrastive learning of natural language processing.

\subsection{Proof of Concept Experiments}

In this section, we discuss the rationale behind designing margins for the two types of augmented positive instances in contrastive learning, from the perspective of mutual information. In the findings of the InfoMin~\cite{tian2020makes} study, they proposed a unified perspective on positive instances for contrastive learning, asserting that reducing the mutual information between positive pairs while preserving task-relevant information is optimal for the task. In the ideal optimization process, the mutual information of positive pairs share only task-relevant information, with no irrelevant noise. Consequently, we build upon their study and demonstrate, both experimentally and theoretically, that our designed margins can guide our positive instances to achieve this desired balance. In our IFCL framework, there is a two-stage exploration of the mutual information of instances and the task-relevant information. We firstly examine the mutual information between the dropout augmented instances $h$ and the fusion augmented instances $h^*$, and investigate whether the margins can effectively reduce their mutual information. And then, we explore the task-relevant information of the textual semantic task, and assess whether the shared mutual information predominantly encompasses task-relevant information without extraneous noise.

For the first stage discussion, we follow the study of InfoMin and define the mutual information $\mathbb{MI}$ between the set of two types of instances $\left\{h_i,h^*_i \right\}^{N}_{i=1}$ as follows:
	\begin{equation}
	\begin{aligned}
		\mathbb{MI}(h,h^*)=  \log(N)+\frac{1}{N}{\sum_{i=1}^{N}\log \frac{e^{\operatorname{sim}\left({h}_{i}, {h}_{i}^{*}\right) / \tau}}
			{{\sum_{j=1}^{N} e^{\operatorname{sim}\left({h}_{i}, {h}_{j}^{*}\right) / \tau}}}},
\end{aligned}
\label{mi}
\end{equation}
In the first stage, we calculate the mutual information of our identical twins $\mathbb{MI}(h,h^+)$ and fraternal twins $\mathbb{MI}(h,h^-)$ after training as shown in Table~\ref{mutual}. We find that both mutual information of our two positive pairs decline in our IFCL, which proves that the margins in \textit{Twins Loss} can reduce the mutual information. Comparing between $\mathbb{MI}(h,h^+)$ and $\mathbb{MI}(h,h^-)$, the designed margins can keep more diverse semantic information from our fusion augmentation in fraternal twins, where the $\mathbb{MI}(h,h^-)$ is higher than $\mathbb{MI}(h,h^+)$ in IFCL.
\begin{table}[!t]
	\caption{Mutual information and task-relevant information. The {IFCL-BERT w/o TL} means training IFCL-BERT without Twins Loss. The experiments are conducted with EnData and STS-B datasets on Bert-base.}
	\centering
	\begin{tabular}{lccc}
		\hline
		\textbf{Method} &\textbf{$\mathbb{MI}(h,h^+)$} &\textbf{$\mathbb{MI}(h,h^-)$}& \textbf{$\mathbb{MI}_{task}$}\\
		\hline
		{IFCL-BERT} & 4.15&4.17&4.31\\
		{IFCL-BERT w/o TL}&4.23&4.20&4.58\\
		{SimCSE}&4.24&-&4.52\\
		\hline
		\label{mutual}
	\end{tabular}
	
\end{table}
In the second stage, we explore why such reduced mutual information is effective for the task in terms of task-relevant information. In our semantic textual tasks, we evaluate the semantic information between two sentences. Therefore, we define the mutual of sentence pairs which have the highest similarity score 5.0 in STS-Benchmark dataset as task-relevant information. As shown in Table~\ref{mutual}, the $\mathbb{MI}_{task}$ in our IFCL is the lowest compared with the others and the most closest to train data $\mathbb{MI}(h,h^+)$ and $\mathbb{MI}(h,h^-)$. Refer to InfoMin, the mutual information of positive pairs in our IFCL contains more task-relevant information. It is the core influence of our designed margins for two types of positive pairs, leading to stronger model semantic recognition.

						\section{Conclusion}
						In this paper, we propose the IFCL, an identical and fraternal twins of contrastive learning, which learns the sentence representations in a more fine-grained semantic manner in the embedding space, and which also provides a new perspective on the contrastive learning between positive instances. With a designed \textit{Twins Loss}, our IFCL significantly improves the efficiency and performance of unsupervised sentence representation on semantic textual similarity tasks in both English and Chinese domains. In the future, we will optimize our method for supervised learning, where the labels are fully used for more diverse positive pairs. 
						\section*{ACKNOWLEDGMENTS}
						This work was supported by Natural Science Foundation of Guangdong (2023A1515012073), National Natural Science Foundation of China(No. 62006083), and National Key Research and Development Program of China (2020YFA0712500).
						
%
%
%
%

							\bibliography{ijcai23}

\begin{thebibliography}{10}

\bibitem{sts15}
Eneko Agirre, Carmen Banea, Claire Cardie, Daniel Cer, Mona Diab, Aitor
  Gonzalez-Agirre, Weiwei Guo, Inigo Lopez-Gazpio, Montse Maritxalar, Rada
  Mihalcea, et~al., `Semeval-2015 task 2: Semantic textual similarity, english,
  spanish and pilot on interpretability', in {\em Proceedings of the 9th
  international workshop on semantic evaluation (SemEval 2015)}, pp. 252--263,
  (2015).

\bibitem{sts14}
Eneko Agirre, Carmen Banea, Claire Cardie, Daniel~M Cer, Mona~T Diab, Aitor
  Gonzalez-Agirre, Weiwei Guo, Rada Mihalcea, German Rigau, and Janyce Wiebe,
  `Semeval-2014 task 10: Multilingual semantic textual similarity.', in {\em
  SemEval@ COLING}, pp. 81--91, (2014).

\bibitem{sts16}
Eneko Agirre, Carmen Banea, Daniel Cer, Mona Diab, Aitor Gonzalez~Agirre, Rada
  Mihalcea, German Rigau~Claramunt, and Janyce Wiebe, `Semeval-2016 task 1:
  Semantic textual similarity, monolingual and cross-lingual evaluation', in
  {\em SemEval-2016. 10th International Workshop on Semantic Evaluation; 2016
  Jun 16-17; San Diego, CA. Stroudsburg (PA): ACL; 2016. p. 497-511.} ACL
  (Association for Computational Linguistics), (2016).

\bibitem{sts12}
Eneko Agirre, Daniel Cer, Mona Diab, and Aitor Gonzalez-Agirre, `Semeval-2012
  task 6: A pilot on semantic textual similarity', in {\em * SEM 2012: The
  First Joint Conference on Lexical and Computational Semantics--Volume 1:
  Proceedings of the main conference and the shared task, and Volume 2:
  Proceedings of the Sixth International Workshop on Semantic Evaluation
  (SemEval 2012)}, pp. 385--393, (2012).

\bibitem{sts13}
Eneko Agirre, Daniel Cer, Mona Diab, Aitor Gonzalez-Agirre, and Weiwei Guo, `*
  sem 2013 shared task: Semantic textual similarity', in {\em Second joint
  conference on lexical and computational semantics (* SEM), volume 1:
  proceedings of the Main conference and the shared task: semantic textual
  similarity}, pp. 32--43, (2013).

\bibitem{mocose2022}
Rui Cao, Yihao Wang, Yuxin Liang, Ling Gao, Jie Zheng, Jie Ren, and Zheng Wang,
  `Exploring the impact of negative samples of contrastive learning: A case
  study of sentence embeddin', {\em arXiv preprint arXiv:2202.13093}, (2022).

\bibitem{stsb}
Daniel Cer, Mona Diab, Eneko Agirre, Inigo Lopez-Gazpio, and Lucia Specia,
  `Semeval-2017 task 1: Semantic textual similarity-multilingual and
  cross-lingual focused evaluation', {\em arXiv preprint arXiv:1708.00055},
  (2017).

\bibitem{use2018}
Daniel Cer, Yinfei Yang, Sheng-yi Kong, Nan Hua, Nicole Limtiaco, Rhomni~St
  John, Noah Constant, Mario Guajardo-Cespedes, Steve Yuan, Chris Tar, et~al.,
  `Universal sentence encoder', {\em arXiv preprint arXiv:1803.11175}, (2018).

\bibitem{simclr2020}
Ting Chen, Simon Kornblith, Mohammad Norouzi, and Geoffrey Hinton, `A simple
  framework for contrastive learning of visual representations', in {\em
  International conference on machine learning}, pp. 1597--1607. PMLR, (2020).

\bibitem{diffcse2022}
Yung-Sung Chuang, Rumen Dangovski, Hongyin Luo, Yang Zhang, Shiyu Chang, Marin
  Solja{\v{c}}i{\'c}, Shang-Wen Li, Wen-tau Yih, Yoon Kim, and James Glass,
  `Diffcse: Difference-based contrastive learning for sentence embeddings',
  {\em arXiv preprint arXiv:2204.10298}, (2022).

\bibitem{infersent2017}
Alexis Conneau, Douwe Kiela, Holger Schwenk, Loic Barrault, and Antoine Bordes,
  `Supervised learning of universal sentence representations from natural
  language inference data', {\em arXiv preprint arXiv:1705.02364}, (2017).

\bibitem{bert2018}
Jacob Devlin, Ming-Wei Chang, Kenton Lee, and Kristina Toutanova, `Bert:
  Pre-training of deep bidirectional transformers for language understanding',
  {\em arXiv preprint arXiv:1810.04805}, (2018).

\bibitem{cert2020}
Hongchao Fang, Sicheng Wang, Meng Zhou, Jiayuan Ding, and Pengtao Xie, `Cert:
  Contrastive self-supervised learning for language understanding', {\em arXiv
  preprint arXiv:2005.12766}, (2020).

\bibitem{simcse2021}
Tianyu Gao, Xingcheng Yao, and Danqi Chen, `Simcse: Simple contrastive learning
  of sentence embeddings', {\em arXiv preprint arXiv:2104.08821}, (2021).

\bibitem{moco2020}
Kaiming He, Haoqi Fan, Yuxin Wu, Saining Xie, and Ross Girshick, `Momentum
  contrast for unsupervised visual representation learning', in {\em
  Proceedings of the IEEE/CVF conference on computer vision and pattern
  recognition}, pp. 9729--9738, (2020).

\bibitem{cnn2014}
Baotian Hu, Zhengdong Lu, Hang Li, and Qingcai Chen, `Convolutional neural
  network architectures for matching natural language sentences', {\em Advances
  in neural information processing systems}, {\bf 27}, (2014).

\bibitem{skipthought2015}
Ryan Kiros, Yukun Zhu, Russ~R Salakhutdinov, Richard Zemel, Raquel Urtasun,
  Antonio Torralba, and Sanja Fidler, `Skip-thought vectors', {\em Advances in
  neural information processing systems}, {\bf 28}, (2015).

\bibitem{bert-flow2020}
Bohan Li, Hao Zhou, Junxian He, Mingxuan Wang, Yiming Yang, and Lei Li, `On the
  sentence embeddings from pre-trained language models', {\em arXiv preprint
  arXiv:2011.05864}, (2020).

\bibitem{libovicky2019language}
Jind{\v{r}}ich Libovick{\`y}, Rudolf Rosa, and Alexander Fraser, `How
  language-neutral is multilingual bert?', {\em arXiv preprint
  arXiv:1911.03310}, (2019).

\bibitem{sickr}
Marco Marelli, Stefano Menini, Marco Baroni, Luisa Bentivogli, Raffaella
  Bernardi, and Roberto Zamparelli, `A sick cure for the evaluation of
  compositional distributional semantic models', in {\em Proceedings of the
  Ninth International Conference on Language Resources and Evaluation
  (LREC'14)}, pp. 216--223, (2014).

\bibitem{word2vec2013}
Tomas Mikolov, Ilya Sutskever, Kai Chen, Greg~S Corrado, and Jeff Dean,
  `Distributed representations of words and phrases and their
  compositionality', {\em Advances in neural information processing systems},
  {\bf 26}, (2013).

\bibitem{glove2014}
Jeffrey Pennington, Richard Socher, and Christopher~D Manning, `Glove: Global
  vectors for word representation', in {\em Proceedings of the 2014 conference
  on empirical methods in natural language processing (EMNLP)}, pp. 1532--1543,
  (2014).

\bibitem{sbert2019}
Nils Reimers and Iryna Gurevych, `Sentence-bert: Sentence embeddings using
  siamese bert-networks', {\em arXiv preprint arXiv:1908.10084}, (2019).

\bibitem{idf2004}
Stephen Robertson, `Understanding inverse document frequency: on theoretical
  arguments for idf', {\em Journal of documentation}, (2004).

\bibitem{cstsb}
Tang Shancheng, Bai Yunyue, and Ma~Fuyu, `A semantic text similarity model for
  double short chinese sequences', in {\em 2018 International Conference on
  Intelligent Transportation, Big Data \& Smart City (ICITBS)}, pp. 736--739.
  IEEE, (2018).

\bibitem{bert-whitening2021}
Jianlin Su, Jiarun Cao, Weijie Liu, and Yangyiwen Ou, `Whitening sentence
  representations for better semantics and faster retrieval', {\em arXiv
  preprint arXiv:2103.15316}, (2021).

\bibitem{pt2022}
Haochen Tan, Wei Shao, Han Wu, Ke~Yang, and Linqi Song, `A sentence is worth
  128 pseudo tokens: A semantic-aware contrastive learning framework for
  sentence embeddings', {\em arXiv preprint arXiv:2203.05877}, (2022).

\bibitem{tian2020makes}
Yonglong Tian, Chen Sun, Ben Poole, Dilip Krishnan, Cordelia Schmid, and
  Phillip Isola, `What makes for good views for contrastive learning?', {\em
  Advances in neural information processing systems}, {\bf 33},  6827--6839,
  (2020).

\bibitem{sncse2022}
Hao Wang, Yangguang Li, Zhen Huang, Yong Dou, Lingpeng Kong, and Jing Shao,
  `Sncse: Contrastive learning for unsupervised sentence embedding with soft
  negative samples', {\em arXiv preprint arXiv:2201.05979}, (2022).

\bibitem{translation2017}
John Wieting and Kevin Gimpel, `Paranmt-50m: Pushing the limits of paraphrastic
  sentence embeddings with millions of machine translations', {\em arXiv
  preprint arXiv:1711.05732}, (2017).

\bibitem{simclue}
Liang Xu, Hai Hu, Xuanwei Zhang, Lu~Li, Chenjie Cao, Yudong Li, Yechen Xu, Kai
  Sun, Dian Yu, Cong Yu, et~al., `Clue: A chinese language understanding
  evaluation benchmark', {\em arXiv preprint arXiv:2004.05986}, (2020).

\bibitem{consert2021}
Yuanmeng Yan, Rumei Li, Sirui Wang, Fuzheng Zhang, Wei Wu, and Weiran Xu,
  `Consert: A contrastive framework for self-supervised sentence representation
  transfer', {\em arXiv preprint arXiv:2105.11741}, (2021).

\bibitem{vascl2021}
Dejiao Zhang, Wei Xiao, Henghui Zhu, Xiaofei Ma, and Andrew~O Arnold, `Virtual
  augmentation supported contrastive learning of sentence representations',
  {\em arXiv preprint arXiv:2110.08552}, (2021).

\bibitem{arccse2022}
Yuhao Zhang, Hongji Zhu, Yongliang Wang, Nan Xu, Xiaobo Li, and Binqiang Zhao,
  `A contrastive framework for learning sentence representations from pairwise
  and triple-wise perspective in angular space', in {\em Proceedings of the
  60th Annual Meeting of the Association for Computational Linguistics (Volume
  1: Long Papers)}, pp. 4892--4903, (2022).

\bibitem{dclr2022}
Kun Zhou, Beichen Zhang, Wayne~Xin Zhao, and Ji-Rong Wen, `Debiased contrastive
  learning of unsupervised sentence representations', {\em arXiv preprint
  arXiv:2205.00656}, (2022).

\end{thebibliography}
							\appendix
							\section{Algorithm}
							\label{sec:algorithm}
							We present the detail algorithm for our IFCL training as follows:
							\begin{algorithm}[ht!] \SetKwData{Left}{left}\SetKwData{This}{this}\SetKwData{Up}{up} \SetKwFunction{Union}{Union}\SetKwFunction{FindCompress}{FindCompress} \SetKwInOut{Input}{input}\SetKwInOut{Output}{output}
								\caption{Training process for each mini-batch in the IFCL.}
								\label{process}
								\Input{A mini-batch sentences $\left \{x_{i}\right \}^{N}_{i=1}$ and the hippocampus queue mechanism $\left \{H_{m},p_{m}\right \}^{k*N}_{m=1}$.} 
								\Output{Optimize the parameters of the IFCL.}
								\emph{\#Generate three types of sentence representations for contrastive learning.}\label{algo}\\
								\For{$i\leftarrow 1$ \KwTo $N$}{ 
									{	${emb}_{i}=Dropout(Embedding(x_{i}),{z}_{i})$\;
										${emb}_{i}^{+}=Dropout(Embedding(x_{i}),{z}_{i}^{+})$\;
										${x}_{i}^{-} =Translate(x_{i})$\;
										${emb}_{i}^{-}=Dropout(Fusion(Embedding(x_{i}),Embedding(x_{i}^{-})),{z}_{i}^{-})$\;
										${h}_{i}=Encode({emb}_{i})$\;
										${h}_{i}^{+}= Encode({emb}_{i}^{+})$\;
										${h}_{i}^{-} = Encode({emb}_{i}^{-})$\;		
								}}
								\emph{\#Calculate the total loss combined with Eq.~(\ref{5}),~(\ref{6}),~(\ref{centerloss}).}\\
								$loss = 0$\;
								\For{$i\leftarrow 1$ \KwTo $N$}{
									$A, B, C=0$\;
									\For{$m\leftarrow 1$ \KwTo $k*N$}{
										$C$ += $p_{m}*Cos\_sim(h_{i},H_{m})$\;
									}
									\For{$j\leftarrow 1$ \KwTo $N$}{
										$A$ += $Cos\_sim(h_{i},h_{j}^{+})$\;
										$B$ += $Cos\_sim(h_{i},h_{j}^{-})$\;
										$\mathbf{M}_j= e^{\operatorname{sim}\left(\mathbf{emb}_{i}, \mathbf{emb}_{j}^{+}\right)}-
										e^{\operatorname{sim}\left(\mathbf{emb}_{i}, \mathbf{emb}_{j}^{-}\right)}$
									}
									\For{$j\leftarrow 1$ \KwTo $N$}{
										$loss$ += $\ell_{i}^{I}(Cos\_sim(h_{i},h_{j}^{+}), A+C)$\
										+$\ell_{i}^{F}(Cos\_sim(h_{i},h_{j}^{-}), B)$\
										+$\ell_{i}^{T}(Cos\_sim(h_{i},h_{j}^{+}), Cos\_sim(h_{i},h_{j}^{-}), \mathbf{M}_j)$
									}
								}
								\emph{\#Update the queue of hippocampus mechanism and the parameters of encoder.}\\
								{
									update $\left \{H_{m}\right \}^{k*N}_{m=1}$ with $\left \{h_{i}\right \}^{N}_{n=1}$\;
									$optimizer(Encode, loss)$\;
								}
							\end{algorithm} 
							
							\section{Influence of Hippocampus Queue Length}
							\begin{table}[h]
								\centering
								\caption{The influence of different hippocampus queue lengths on the STS-B and C-STS-B tasks in the IFCL, where {M=k*N}. }
								\begin{tabular}{lcc}
									\hline
									{Queue length M} & {STS-B} & {C-STS-B}\\
									\hline
									{M=320} &{78.76} & {69.53} \\
									{M=384} & {79.15} & {70.18}\\
									{M=416} &\textbf{80.27}& {70.60}\\
									{M=448}& {79.79} & \textbf{71.41}\\
									{M=512} & {78.13} & {71.08}\\
									\hline
								\end{tabular}

								\label{queue length}
							\end{table}
							In the hippocampus queue mechanism, the more negative instances stored in the queue could benefit to bring in a more diverse variety of features, which is efficient to prevent the collapse of representations space. To verify the effect of the quantity of the negative instances, we test the IFCL with different queue lengths on the STS-B and C-STS-B tasks to investigate the impact of the performances. 
							As shown in Table~\ref{queue length}, the IFCL performs best when {M=416} for STS-B task and {M=448} for C-STS-B task. However, it performs worse as the queue length continuously increases, where more noise would be introduced into the training process.
							
							\section{Implement Details}	
							\label{sec:details}
							We implement the IFCL based on Sentence-BERT \cite{sbert2019} and initialize models with three different versions of pre-trained  BERT$_{base}$ \cite{bert2018} and corresponding language models, including Bert-Base-Uncased\footnote{\url{https://huggingface.co/bert-base-uncased}}  (Bert-Base-Multilingual-Uncased\footnote{\url{https://huggingface.co/bert-base-multilingual-uncased}} ), Roberta-Base\footnote{\url{https://huggingface.co/roberta-base}}  (Bert-Base-Multilingual-Uncased), Bert-Base-Chinese\footnote{\url{https://huggingface.co/bert-base-chinese}}  (Denpa92/Bert-Base-Cantonese\footnote{\url{https://huggingface.co/denpa92/bert-base-cantonese}} ). For the fraternal twins, we generate the fraternal instances on EnData by feeding the German to the framework together with English, and also, generate the fraternal instances on CnData by feeding the Cantonese sentences to the framework together with Chinese. We use an NVIDIA GeForce RTX 2080Ti (11g) GPU to fine-tune our models, which limits the batch size to 64. In our experiments, we fine-tune the IFCL for 1 epoch and evaluate the models with the verification sets of STS-B or C-STST-B \cite{stsb,cstsb} every 151 steps. We choose Spearman's correlation as the evaluation metric, which is the same as the previous works. The hyperparameters in our implements are shown in Table~\ref{hyperparameters}.
							\footnote{\url{https://huggingface.co/denpa92/bert-base-cantonese}} 
							\footnote{\url{https://huggingface.co/peterchou/simbert-chinese-base}}
							\begin{table}[ht]
								\setlength\tabcolsep{10pt}
								\centering
								\begin{tabular}{lr r}
										\hline
										{Hyperparameters} & {EnData} & {CnData}\\
										\hline
										{Learning rate} &{1e-5} & {1e-5}\\
										{Dropout rate $\rho$} &{0.15} & {0.2} \\
										{Fusion rate $\varepsilon$} & {0.9} & {0.3}\\
										{Forgetting rate $\lambda$} &{0.002}& {0.003}\\
										{Temperature $\tau$}& {0.05} & {0.034}\\
										{Similarity $\omega_{a}$} & {0.85} & {0.85}\\
										{Similarity $\omega_{b}$} & {0.80} & {0.80}\\
										{Max sequence length}&{32}&{64}\\
										{Batch size N} &{64}&{64}\\
										{Queue length k*N} &{416}&{448}\\
										\hline
									\end{tabular}
								\caption{Hyperparemeters for training IFCL in EnData and CnData datasets. }
								\label{hyperparameters}
							\end{table}
							
							\section{Discussion}
							\label{sec:discussion}
							Based on the above analysis, we explain the role of each important component in the IFCL. From the perspective of the positive instances, the fusion augmentation not only increases the quantity of the positive instances but also enriches the variety of the positive pairs. Therefore, the IFCL breaks the limits with comparisons only between the positive and negative instances but pays more attention to the diversity of the positive instances. On the other side, the hippocampus queue mechanism fully utilizes the datasets and computational resources for contrastive learning, which improves the performance of the IFCL and reduces the number of datasets required.
							
						\end{document}